\pdfoutput=1
\documentclass[11pt]{article}
\usepackage{natbib}

\usepackage{EMNLP2022}

\usepackage{times}
\usepackage{latexsym}

\usepackage[T1]{fontenc}

\usepackage[utf8]{inputenc}

\usepackage{inconsolata}

\usepackage{hyperref}
\usepackage{url}
\usepackage{graphicx}
\usepackage{multirow}
\usepackage{nicematrix}
\usepackage{booktabs}
\usepackage{amsfonts}
\usepackage{subcaption}

\title{Domain Curricula for Code-Switched MT at MixMT 2022}

\author{Lekan Raheem \& Maab Elrashid \\
         African Institute for Mathematical Sciences (AIMS) \\ 
         \texttt{\{rwaliyu,mnimir\}@aimsammi.org}}
\date{}

\begin{document}

\maketitle

\begin{abstract}
In multilingual colloquial settings, it is a habitual occurrence to compose expressions of text or speech containing tokens or phrases of different languages, a phenomenon popularly known as code-switching or code-mixing (CMX). We present our approach and results for the Code-mixed Machine Translation (MixMT) shared task at WMT 2022: the task consists of two subtasks, monolingual to code-mixed machine translation (Subtask-1) and code-mixed to monolingual machine translation (Subtask-2). Most non-synthetic code-mixed data are from social media but gathering a significant amount of this kind of data would be laborious and this form of data has more writing variation than other domains, so for both subtasks, we experimented with data schedules for out-of-domain data. We jointly learn multiple domains of text by pretraining and fine-tuning, combined with a sentence alignment objective. We found that switching between domains caused improved performance in the domains seen earliest during training,  but depleted the performance on the remaining domains.  A continuous  training run with strategically dispensed data of different domains showed a significantly improved performance over fine-tuning.
\end{abstract}
\section{Introduction}
Code-mixing (CMX) denotes the alternation of two languages within a single utterance \cite{articlePoplack, DBLP:journals/corr/abs-1904-00784}. Code-mixing occurs mostly in unofficial groups in multilingual environments. More than 77\% of Asians are multilingual \cite{articlex}, and other statistics estimate that 64.5\% of Europeans speak more than two languages, with more than 80\% of adults in the region being bilingual \cite{eurostats}. Code-mixing happens far more often in conversations than in writing, and mostly in unofficial settings, hence it rarely occurs in documented settings. This makes  substantial data gathering for computational approaches to translations of code-mixed language difficult.  Parallel corpora for code-switched data is very scarce \cite{menacer:hal-02106010}, this is because code-mixing mostly occurs in unofficial conversations like social media interactions.

Contemporary Neural Machine Translation (NMT) mostly makes use of parametric sequence-to-sequence models \cite{https://doi.org/10.48550/arxiv.1409.0473, DBLP:journals/corr/VaswaniSPUJGKP17}, where an encoder receives a source sentence and outputs a set of hidden states, the decoder then scrutinizes these hidden states at each step, and outputs a sequence of softmax distribution over the target vocabulary space. Considering that we would need vast quantities of data to train an adequate NMT for this task, we leverage large-scale synthetic and available small data and notably rank data on domain relevance, by fine-tuning with it, initiating training with the relevant domain and strategically placing it at the premier batches of the training data. 

Essentially, the characteristics of the data an NMT model is trained on are paramount to its translation quality, in particular in terms of size and domain. It is quintessential to train NMT models based on the domain relevance of corpora.  Since most code-mixing occurs in unofficial communication, it is costly to find a lot of labeled data for every domain we are interested in.  Hence we attempt to find less expensive exigencies to supplement training data, pretrain on largely available data of different domains, strategically construct synthetic data, and apportion data to make up for missing domains.

In these WMT Subtasks – monolingual to code-mixed machine translation (Subtask-1) and code-mixed to monolingual machine translation (Subtask-2), we also fine-tune on different domains, align representations of data and find the best combination of approaches to solving the subtasks. The main intuition behind our proposed solution is that NMT models exhibit a significant translation correlation when trained on data from the same or similar domains. With different data domain requirements, it performs better when trained with data of the most relevant domain as preliminary batches compared to finetuning. As most natural code-mixed data source is social media and it is difficult to gather a good amount to train a model, it is incumbent to find a strategy that makes the model prioritize this form of data above others. Accordingly, we attempt to find less expensive techniques to supplement training data, pretrain on largely available data of different domains, strategically construct synthetic data, and apportion data to make up for missing domains. Our result showed improved performance on innate code-mixed data (and non-synthetic WMT test set samples) when this was prioritized and performed strongly in a test with a mix of several other data sources. We observed a better performance with domain-specific evaluation upon finetuning but this intensely plummeted performance on other ‘pretraining domains’, and more balanced performance on passing the interesting domain in the preliminary batches in a single ‘all domain’ training.

\section{Related Work}
It is laborious to obtain ‘one-fits-all’ training data for NMT. Most publicly available parallel corpora like Tanzil, OPUS, UNPC are sourced from documented communication, and these are often domain-specific. In NMT, data selection e.g. \citet{axelrod-etal-2011-domain} has remained as an underlying and important research concern. Choosing training examples that are relevant to the target domain, or by choosing high-quality examples for data cleaning (also known as denoising), has been essential in domain adaptation. Building a large-scale multi-domain NMT model that excels on several domains simultaneously becomes both technically difficult and practically back-breaking. Addressing research problems such as catastrophic forgetting \cite{https://doi.org/10.48550/arxiv.1312.6211}, data balancing \cite{DBLP:journals/corr/abs-2004-06748}, Adapters \cite{pmlr-v97-houlsby19a} have shown improvement. Unfortunately, several domains are difficult to handle with the single-domain data selection techniques currently in use. For instance, improving translation quality of one domain will often hurt that of another \cite{britz-etal-2017-effective, van-der-wees-etal-2017-dynamic}.

\citet{DBLP:journals/corr/abs-1904-09107} replaced phrases with pre-specified translation to perform “soft” constraint decoding. \citet{DBLP:journals/corr/abs-2105-04846} generated code-mixed data from regular parallel texts and showed this training strategy yields MT systems that surpass multilingual systems for code-mixed texts. 

Considering that code-mixed text belongs in less documented domains than most, there may be a need for domain adaptation used on sufficiently available data domains. Our work is inspired by the following approaches: \citet{DBLP:journals/corr/abs-1908-10940} executed simultaneous data selection across several domains by gradually focusing on multi-domain relevant and noise-reduced data batches while carefully introducing instance-level domain-relevance features and automatically constructing a training curriculum. \citet{https://doi.org/10.48550/arxiv.2204.09259} demonstrated that instance-level features are better able to distinguish between different domains compared to corpus-level attributes. \citet{DBLP:journals/corr/abs-1910-02555} proposed modeling the difference between domains instead of smoothing over domains for machine translation.

\citet{anwar2022true} showed that an encoder alignment objective is beneficial for code-mixed translation, in addition to \citet{DBLP:journals/corr/abs-1903-07091} that proposed auxiliary losses on the NMT encoder that imposed representational invariance across languages for multilingual machine translation.

\begin{table}[htbp]
\centering
 \setlength\tabcolsep{4pt}
\scalebox{0.80}{
\begin{tabular}{p{4.3cm}p{4.3cm}}
 \toprule
 English & Code-Mixed (\textsc{CMX}) \\
 \midrule
@dh*v*l2410*6 sure brother :) &  @dh*v*l2410*6 sure bhai :) \\
  \addlinespace[0.2cm]
"I just need reviews like these, this motivates me a lot" &  "Bas aise hi reviews ki zaroorat hai, kaafi protsahan milta hai in baaton se. " \\
  \addlinespace[0.2cm]
When the sorrow got missing in this room, the blood also became thin, \#GuessTheSong & 
 Jab gam ye rum mein kho gaya, toh khoon bhi patla hogaya \#GuessTheSong \\
  \bottomrule
 \end{tabular}}
\caption{Examples from the WMT Shared Task Dataset.}
\label{tab:example_wmt} 

\end{table}

\begin{table}[htbp]
\centering
 \setlength\tabcolsep{4pt}
\scalebox{0.80}{
\begin{tabular}{p{4.3cm}p{4.3cm}}
 \toprule
 English & Code-Mixed (\textsc{CMX}) \\
 \midrule
Overhead charge is a percentage of the direct costs of providing the services under the contract. &  Overhead charge, anubandh ke anusaar pradatt sevaon kee pratyaksh laagat ka ek pratishat hota hai. \\

  \addlinespace[0.2cm]
A strategy of ignoring potential problems on the basis that they may be exceedingly rare. &  us aadhaar par sambhaavit problems ko anadekha karane kee ek yukti, jahaan ki ve ati dushpraapy ho sakate hain. \\

    \addlinespace[0.2cm]
    A standard of measurement, or a unit that can be studied separately / independently. & koee maapadand athava koee a unit that svatantr roop se/alag se adhyayan kiya ja sakata ho.\\
   \bottomrule
\end{tabular}}
\caption{Examples from IITB Corpus.}
\label{tab:example_iitb} 

\end{table}

\section{Data}
\label{data-section}
In table ~\ref{tab:example_wmt} we show some samples from the WMT shared task, sourced from the non-synthetic validation data. The data provided for Subtask-1 \cite{srivastava2021hinge} contains synthetic and human-generated data and Subtask-2 Parallel Hinglish Social Media Code-Mixed Corpus \cite{srivastava2020phinc}  for both tasks are mostly unofficial, mostly short conversational sentences, with some letters asterisked for privacy/derogatory reasons.

Since we need to augment provided data for a reasonable quantity to train a NMT model, we generated synthetic code-mixed data from the IITB corpus \cite{DBLP:journals/corr/abs-1710-02855} which is from 17 sources of different domain mostly HindEnCorp \cite{bojar-etal-2014-hindencorp}, Gyaan-Nidhi Corpus \cite{articleKamal},  Indian Government corpora - CFILT, Mahashabdkosh, Tanzil, and GNOME \cite{DBLP:journals/corr/abs-1710-02855} (details in section ~\ref{csw}). Synthetic code-switched sentences generated from  the IITB corpus belong to a different domain than the WMT evaluation data, as we illustrate with the English translation samples in table ~\ref{tab:example_iitb}.

For the pretraining-finetuning setup, we pretrain with synthetic code-switched data generated from IITB corpus and fine-tune on the WMT data provided for each task. For both pretraining and fine-tuning, we coordinate the data similar to \cite{anwar2022true} – For Subtask-1, Monolingual to code-mixed machine translation subtask, we use the Hindi sentence (Devanagari script) as source sequence and  the corresponding code-switched sentence (Roman script) as target, then alternated the English sequence (Roman script) as source sentence and the same corresponding code-switched sentence as the target sequence. The above two source-target parallel data are set after each other. For Subtask-2, Code-mixed to monolingual machine translation subtask, we have a similar arrangement as in Subtask-1, but with the source sequences of Hindi and code-mixing (Hinglish) in Roman script and as the target the corresponding English sequence. We removed sequences shorter than 2 tokens, and those longer than 250 tokens, and a target-to-source token ratio of more than 1.5. After cleaning the pretraining data, for Subtask-1, we have about 2.5M parallel sentences and 2.3M parallel sentences for Subtask-2.  

For the finetuning process, we made use of the WMT training data provided for each subtask and organized like the pretraining data as described above. After cleaning, for Subtask-1 (Synthetic + Human-generated), we have a total of over 11K parallel sentences. For subtask-2, over 12K parallel sentences remain after cleaning. 

Since the IITB corpus encompasses multiple sources and domains where code-mixing infrequently occurs, we decided to configure our model in a way it first learns from natural code-mixed data provided by WMT. We experiment with a hand-designed curriculum of the Synthetic Code-switched data generated from the IITB corpus and the WMT provided data. We supply the model the non-synthetic WMT data only in the first few batches in the hope that this would faintly familiarize the model with domain-specific features before it learns from the synthetic code-switched data we generated from other domains. We compare the results of this approach to the above described pretraining-finetuning setup. All data is tokenized and normalized using sentencepiece\footnote{https://github.com/google/sentencepiece}.

\subsection{Code Switched Data Generation}
\label{csw}

Given that most publicly available corpora are monolingual, it is requisite to generate sufficient synthetic code-mixed data for training. Moreover, there have been works on generating synthetic code-mixed data linguistically, there are a few rules theories that are essential. 

The \textbf{Equivalence Constraint Theory} states that intra-sentential code-mixing can only happen where the surface structures of two languages map onto each other, implicitly following both languages' grammatical norms \cite{articlePoplack}. Fundamentally, we can only attempt code-mixing at points where both languages coincide on the parse tree to equivalent phrase structure.

The \textbf{Matrix Language Theory} explains code-mixing by introducing the concept of a "Matrix Language," or base language, into which clusters of the "Embedded Language," or second language, are introduced in such a way that the former sets the grammatical structure of the sentence and the latter "switches-in'' at grammatically correct points of the sentence \cite{myers2001matrix}. The Matrix language has more tokens in the sequence and its rules are designated above the embedded language’s.

\begin{figure*}[th]
\centering
\begin{subfigure}{0.8\columnwidth}
  \centering
  \includegraphics[width=\linewidth]{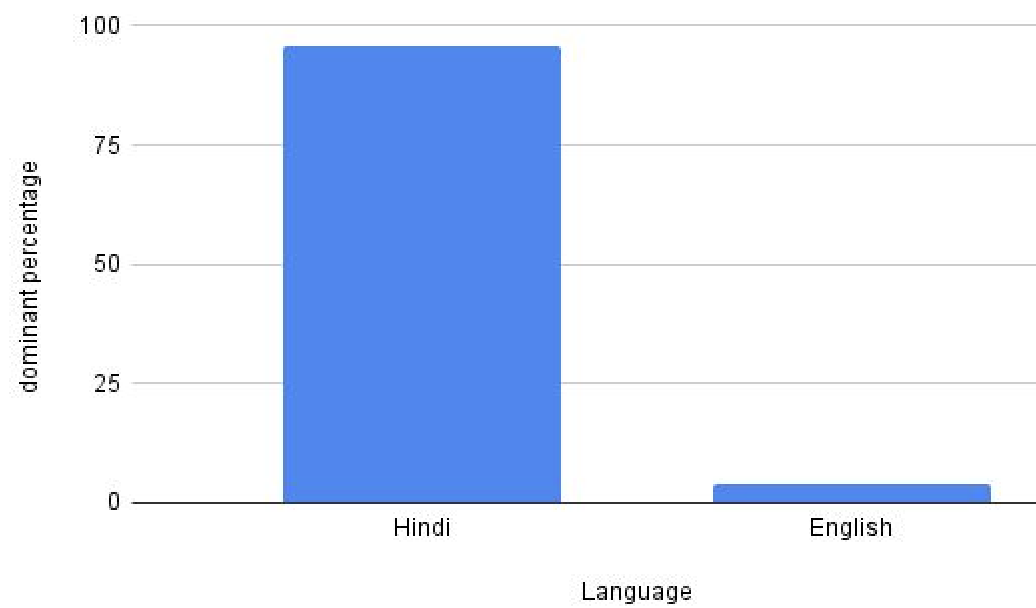}
  \caption{Subtask-1}
  \label{fig:sub1}
\end{subfigure}%
\begin{subfigure}{0.8\columnwidth}
  \centering
  \includegraphics[width=\linewidth]{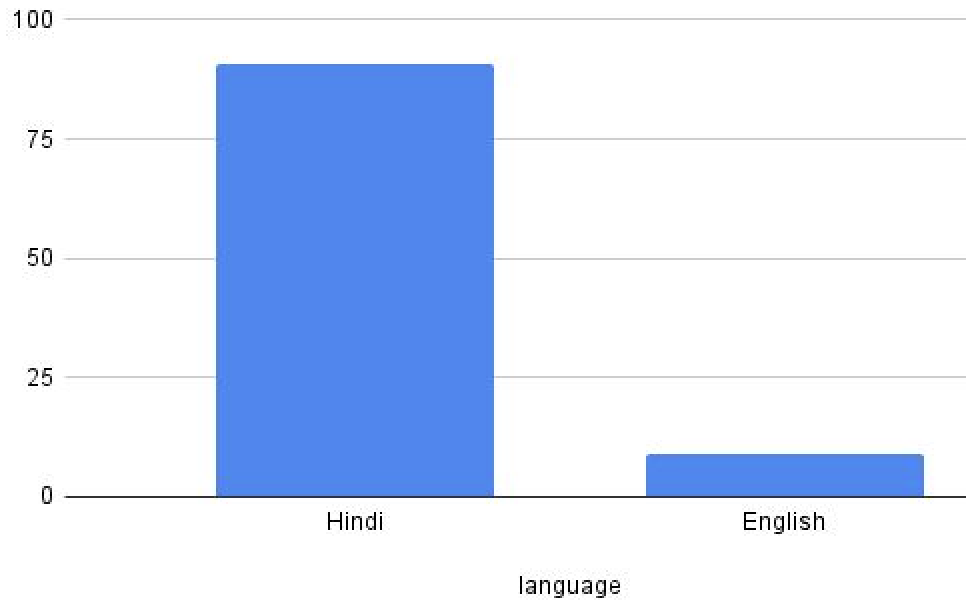}
  \caption{Subtask-2}
  \label{fig:sub2}
\end{subfigure}
\caption{Percentage of Hindi vs. English as matrix language from WMT’22 Hinglish validation data for the subtasks.}
\label{matrixlanguage}
\end{figure*}

Considering the linguistic theories above, we generate code-mixed data by locating where both languages coincide based on a word-level alignment extracted and  only replace tokens based on the “matrix language theory”. Roughly following the recipes by \cite{DBLP:journals/corr/abs-1904-09107, rizvi-etal-2021-gcm, DBLP:journals/corr/abs-2105-04846, anwar2022true}, we generate synthetic code-switched data from the IITB parallel data: We create code-mixed data by first transliterating Hindi (Devanagari script) to Roman script using Ritwik’s tool\footnote{https://github.com/ritwikmishra/devanagari-to-roman-script-transliteration}, then extract word alignments using the giza++ toolkit \cite{och03:asc}, and  extract minimal alignment units following the approach of \cite{articleCrego}. We choose Hindi as the “matrix language” by determining this from the provided WMT training data, we extract word alignments and find how many tokens in each sequence belongs to which language using the language detector of Googletrans python library\footnote{https://github.com/ssut/py-googletrans/blob/master/docs/index.rst} and assign the language with more tokens as the matrix language. Figure ~\ref{matrixlanguage} shows the Hindi/English matrix language ratio for both subtasks.

Similar to MLM pre-training used by BERT \cite{DBLP:journals/corr/abs-1810-04805}, we randomly replace 15\% of the tokens in each Hindi sentence with their aligned segments in the embedded language (English). For short sequences with less than 7 tokens we make only one replacement, chosen uniformly at random.

\section{Training Objective}
Considering the effectiveness of clean finetuning \cite{DBLP:journals/corr/abs-1901-10430}, and pre-training \cite{DBLP:journals/corr/abs-1909-11229}, we attempt a combined pipeline of pretraining$+$finetuning experiment and also a single training but with tactical positioning of the most important domain. In the finetuning process and training with specially ordered data, as recommended by \cite{anwar2022true}, we add an alignment loss to the encoder to encourage source and target representations to be close in representation space minimizing the max-pooled cosine distance of the encoder representation as shown in equation ~\ref{eq:1}:

\begin{equation} 
\label{eq:1}
\Omega = 
\mathbb{E}_{D_{(en,hi)}}[1 - sim(Enc(x_{src}), Enc(x_{tgt}))]
\end{equation}

\begin{figure}[ht]
\centering
\includegraphics[scale=0.25, keepaspectratio]{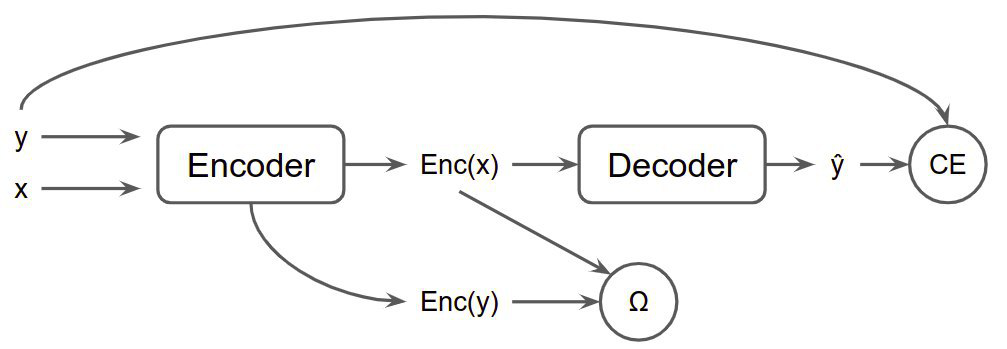}
\caption{The loss function visualization, $CE$ is the Cross Entropy, $\Omega$ is the encoder loss.
}
\label{fig:encoder}
\end{figure}

Where $\Omega$ is the encoder loss, $D_{(en, hi)}$  is the data consisting of the parallel pairs of code-mixed to monolingual or monolingual to code mixed depending on which subtask the data belongs to, $x_{src}$ is the source sequence and $x_{tgt}$ is the target sequence, Enc(x) is the max-pooled encoder representation of sentence x similar to \cite{https://doi.org/10.48550/arxiv.1410.2455} and \cite{coulmance-etal-2015-trans}, and sim is the cosine similarity. Unlike \cite{DBLP:journals/corr/abs-1903-07091} where the whole model’s parameters are updated as shown in figure ~\ref{fig:encoder}.

\section{Experiments and Results}
In all of our experiments, we used Transformer-Base \cite{DBLP:journals/corr/VaswaniSPUJGKP17} configuration with the Fairseq \cite{DBLP:journals/corr/abs-1904-01038} framework. All models were trained on four Tesla T400 GPUs using IITB and WMT-’22 MixMT data for training as described in Section ~\ref{data-section}, with a shared vocabulary of 77K BPE \cite{DBLP:journals/corr/SennrichHB15} sub-words to create a joint vocabulary for both tasks and all models. The model's hyperparameters can be found in Appendix ~\ref{AppendixA}.

\begin{table*}[htbp]
\centering
 \setlength\tabcolsep{4pt}
\scalebox{0.80}{
\begin{tabular}{lrrrrrr}
 \toprule
 \textbf{Model} & \multicolumn{2}{c}{\bf{IITB Eval Set}} & \multicolumn{2}{c}{\bf{WMT Eval Set}} & \multicolumn{2}{c}{\bf{Mixed Eval Set}} \\
& Subtask-1 & Subtask-2 & Subtask-1 & Subtask-2 & Subtask-1 & Subtask-2  \\
\midrule
Pretrained (IITB corpus only) & \bf{0.81} & \bf{0.85} & 0.41 & 0.47 & \bf{0.76} & \bf{0.80}\\
Pretrained (IITB corpus) + Finetuned (WMT provided) &  0.49 & 0.52 & 0.54 & 0.58 & 0.53 & 0.59\\
Mixed-data training (target domain first) & 0.76 & 0.79 & \bf{0.62} & \bf{0.64} & 0.70 & 0.73\\
  \bottomrule
 \end{tabular}
 }
\caption{Translation accuracy of subtask-1 and subtask-2 of Hindi-English in ROUGE-L (F1-Score) on different \textbf{test} data of different domains, based on models trained on different domain training data, data arrangement or training pipeline.}
\label{tab:main_results} 

\end{table*}

\subsection{Results}
Based on the human evaluation  by the organizers of the subtasks, the translation result of our initial models - v0.2 submitted  – which was trained with mixing the IITB with the WMT without prioritizing the target domain – had an overall rating of 1.75 from 10 random translations for each subtask, this ranked inferior to many other submissions. 

With the  help of native Hindi speakers to investigate our data, we found some of the causes it performed decumbent, which were as a result of some of the different data preprocessing tools we used:
For Transliteration, We tried a few devanagari to roman tools but had some shortcomings like:
\begin{itemize}
    \item Lipika-ime\footnote{https://github.com/ratreya/lipika-ime}: inappropriate handling of diacritic characters.
    \item Indic-trans\footnote{https://github.com/libindic/indic-trans}: Removal of vowels (e.g. default -> difolt, highlight -> hilite, method -> methd, etc..), Splitting of words that lead to suboptimal outputs (e.g. "un he" instead of "unhe").
    \item Sheental\footnote{https://github.com/sheetalgiri/devanagari-to-roman-script}: repetition of vowels e.g. jane -> jaane, yaar -> yaara, incorrect replacement of characters e.g. om -> on and occurence of needless suffixes e.g. palat -> palata, some diacritic appeared independently.
    \item Ritwik's: inappropriately breaking very long sentences into multiple lines, replacing individually occurring tokens like um -> oon and abruptly stopping when ran over large amount of data so we divided the data into chunks each containing not more than 200K sequences, optimized by parallel computing using dask, and replaced the individually occurring tokens changed afterwards.
\end{itemize}
    We also investigated our initial model and discovered a few other issues like: 
\begin{itemize}   
    \item Cases of translation of proper nouns in Subtask-2 (e.g. Sapna -> dream) which we deduce as a pointer to insufficient training data.
    \item Imprecise tokenization and detokenization, we also switched to use of Google sentencepiece instead of Moses SMT
    \item Also, the organizers noticed the team's output had an incorrect order. A problem where the post-processing had sorted the hypothesis and fragmented longer sentences also influenced the rating. 
\end{itemize}

Upon inspecting our model outputs we found a few inaccuracies with the tools we used for transliteration and tokenization for the submitted model hypotheses. We fixed these, and present the results in the following section.

\subsection{Post-Submission Results}
\label{post}
Table ~\ref{tab:main_results} shows the experimental results based on different test data of  samples each from IITB corpus, WMT, and a Mixed test sample evenly selected from Samanatar (includes IITB corpus, CCMatrix, Hindi-News, Jagran, Livehindustan, Patrika and WMT). We made use of other reputable tools to fix the aforementioned errors, added the domain curriculum technique, and ran the experiment again and present it in table ~\ref{tab:main_results}.

Table \ref{tab:main_results} shows that fine-tuning on the WMT domain improves translation accuracy on this domain slightly, but the model suffers ‘catastrophic forgetting’ on domains it was initially trained on. Pretraining did not lead to a good generalization for the WMT test samples provided, hence a need for domain adaptation. Placing the relevant domain in the preliminary batches for mixed-data training also improves training on such a domain but hurts other domains slightly.

\section{Conclusion}
We present a data domain sorting method that improves translation performance based on a target domain for the WMT 2022 code-switching shared tasks. We compared our result to a pretraining and fine-tuning pipeline, and demonstrated that the finetuning method improves on specified domain but upsets on previously learned data domain. An aspect we intend to delve further into is efficient domain adaptation strategies that may help low-resource domains such as code-mixing, and have little or no effect on high-resource domains, we are currently looking into domain adaptation learning curve \cite{https://doi.org/10.48550/arxiv.2204.09259}, extraction of domain-specific parameters \cite{DBLP:journals/corr/abs-1910-02555} for better data augmentation strategies, better acquisition of code-mixed data, and the use of Adapters \cite{pmlr-v97-houlsby19a}.

\section*{Acknowledgement}
We would like to express special gratitude to Julia Kreutzer and Melvin Johnson of Google Research, Mohamed Anwar and Sweta Agrawal for their support and contributions. We also appreciate the GCP credits granted by Google for this project.

\bibliography{emnlp2022}
\bibliographystyle{acl_natbib}

\appendix
\section{Appendix}
\label{AppendixA}
Table ~\ref{tab:model_table} holds all the hyper-parameters we used for training all models. All experiments were set to halt at patience of 15 updates on the BLEU \cite{articlePapineni} stabilizing, we found it trained longer with BLEU, but evaluated on WMT specified F1-Score \cite{inproceedingsSokolova} for the subtasks.

\begin{table}[!htb]
\centering
 \setlength\tabcolsep{4pt}
\scalebox{0.80}{
\begin{tabular}{p{4.3cm}p{4.3cm}}
 \toprule
 Hyper-parameter & Value \\
 \midrule
    Number of Layers & 6\\
    Hidden size & 512\\
    FFN inner hidden size & 2048\\
    Attention heads & 8\\
    Attention head size & 64\\
    Dropout & 0.1\\
    Attention Dropout & 0.0\\
    Warmup Steps & 4000\\
    Learning Rate & 5e-4\\
    Learning Rate Decay & inverse\_sqrt\\
    Batch Size & 4096 tokens\\
    Label Smoothing & 0.1\\
    Weight Decay & 0.0001\\
    Adam $\epsilon$ & $10^{-9}$\\
    Adam $\beta_1$ & 0.9\\
    Adam $\beta_2$ & 0.98\\
    Encoder Criterion Weight & 10\\
  \bottomrule
 \end{tabular}}
\caption{The hyperparameter values setting for training.}
\label{tab:model_table} 

\end{table}


\end{document}